\documentclass[letterpaper]{article} 
\usepackage[draft]{aaai25}  
\usepackage{times}  
\usepackage{helvet}  
\usepackage{courier}  
\usepackage[hyphens]{url}  
\usepackage{graphicx} 
\usepackage{natbib}  
\usepackage{caption} 
\usepackage{algorithm}
\usepackage{algorithmic}

\usepackage{newfloat}
\usepackage{listings}

\usepackage{newfloat}
\usepackage{listings}

\usepackage{caption}
\usepackage{makecell}
\usepackage{subcaption}

\usepackage{amsmath,amsfonts,graphicx,bm,amssymb,bbding}
\usepackage{algorithmic}
\usepackage{booktabs}
\usepackage{multirow}
\usepackage{url}

\usepackage{tabularx}
\usepackage{longtable}

\usepackage{float}

\usepackage{color}

\usepackage{endnotes}

\DeclareCaptionStyle{ruled}{labelfont=normalfont,labelsep=colon,strut=off} 
\floatstyle{ruled}
\newfloat{listing}{tb}{lst}{}
\floatname{listing}{Listing}
\title{Enhancing Long Video Understanding via Hierarchical Event-Based Memory}

\author{
    Dingxin Cheng\textsuperscript{\rm 1,2}\thanks{Work done during internship at Tencent, \textsuperscript{$\dagger$}~Corresponding Author.}, Mingda Li\textsuperscript{\rm 1, $\dagger$}, Jingyu Liu\textsuperscript{\rm 1}, Yongxin Guo\textsuperscript{\rm 3},\\ Bin Jiang\textsuperscript{\rm 2, $\dagger$}, Qingbin Liu\textsuperscript{\rm 1}, Xi Chen\textsuperscript{\rm 1}, Bo Zhao\textsuperscript{\rm 1}
}
\affiliations{
    \textsuperscript{\rm 1} Tencent PCG
    \textsuperscript{\rm 2} Shandong University\\
    \textsuperscript{\rm 3} School of Science and Engineering, The Chinese University of Hong Kong (Shenzhen)}

\usepackage{bibentry}

\begin{document}

\maketitle

\begin{abstract}
Recently, integrating visual foundation models into large language models (LLMs) to form video understanding systems has attracted widespread attention. Most of the existing models compress diverse semantic information within the whole video and feed it into LLMs for content comprehension. While this method excels in short video understanding, it may result in a blend of multiple event information in long videos due to coarse compression, which causes information redundancy. Consequently, the semantics of key events might be obscured within the vast information that hinders the model's understanding capabilities. To address this issue, we propose a Hierarchical Event-based Memory-enhanced LLM (HEM-LLM) for better understanding of long videos. Firstly, we design a novel adaptive sequence segmentation scheme to divide multiple events within long videos. In this way, we can perform individual memory modeling for each event to establish intra-event contextual connections, thereby reducing information redundancy. Secondly, while modeling current event, we compress and inject the information of the previous event to enhance the long-term inter-event dependencies in videos. Finally, we perform extensive experiments on various video understanding tasks and the results show that our model achieves state-of-the-art performances.
\end{abstract}
\vspace{-0.2cm}
%

\section{Introduction}
The emergence of Large Language Models (LLMs)~\cite{llama,vicuna} has brought about revolutionary changes in the field of NLP, with their exceptional understanding and reasoning abilities enabling the generation of high-quality language texts across various domains. Nonetheless, to genuinely realize the universality of the model, it needs to be capable of integrating and understanding data stemming from multiple modalities, including images, videos, and audio. In response to this requirement, some researchers aim to harness the potent capabilities of LLMs to concurrently integrate information from multiple modalities, thereby addressing a variety of multimodal tasks~\cite{mplug-2,mplug-owl2}. For instance, video understanding~\cite{vis4mer,trans4mer} serves as a prime example of those multimodal tasks.

To address video understanding tasks, most existing models primarily employ visual foundation models~\cite{evavit} to extract visual tokens, which are then directly fed into LLMs to generate inferential outcomes. However, this can be problematic in that some methods~\cite{video-llava,videochat} only input a smaller number of frames into LLMs for video comprehension due to the limitations of the LLMs context input length and computational resources. While this practice yields satisfactory results for short videos, it may lead to information loss for longer videos, proving detrimental to the temporal modeling of the video content. To cope with this, some methods~\cite{ma-lmm,moviechat,moviechat+} employ token compression to simultaneously process a larger number of frames in order to compensate for information loss. However, in longer videos, such coarse compression may lead to the amalgamation of various event information, resulting in information redundancy. Consequently, the semantics of key events may be obscured by the overwhelming amount of information, thereby affecting the model's comprehension capabilities.

To address the aforementioned challenges, we propose a Hierarchical Event-based Memory-enhanced LLM (HEM-LLM) for better understanding of long videos. Considering the diverse event information contained in long videos, it is crucial that the model processes each event individually to prevent information redundancy. Specifically, we first devise a novel adaptive sequence segmentation scheme to partition multiple events in long videos. In this way, our model can treat each event individually, thereby reducing information clutter. Secondly, we introduce event-based local memory to model individual events, storing information from historical frames within an event to establish intra-event contextual connections. Thirdly, while modeling the current event, we employ global video memory to compress and inject information from the previous event, so enhancing the long-term inter-event dependencies in long videos. Finally, we conduct extensive experiments on various video understanding tasks, such as video question answering, video captioning, and long video activity classification. The results on nine benchmark datasets demonstrate the effectiveness of our model.


\section{Related Work}

\subsection{Multi-modal Large Language Models}
Large language models (LLMs)~\cite{vicuna,gpt3,gpt4} have exhibited remarkable capabilities in language comprehension and reasoning recently, enabling them to generate superior natural language text across various domains. Hence, several researchers have attempted to exploit the potential of LLMs to tackle multimodal tasks by integrating foundational models from other modalities with LLMs, thereby constructing MLLMs~\cite{videollama2,llama-vid,llava-next,vtg-llm} and endowing them with multimodal comprehension capabilities. In the image field, Flamingo~\cite{flamingo} integrates perceiver resampler and gated cross-attention layers to establish a connection between the frozen image encoder and the LLMs. BLIP-2~\cite{blip-2} introduces a versatile and efficient pre-training approach, which initiates vision-language pre-training by leveraging a frozen image foundation encoder and a frozen large language model. LLaVA~\cite{llava} utilizes a straightforward linear layer to map image features into the text embedding space and effectively finetunes LLMs using extensive data to enhance performance. In the video field, VideoChat~\cite{videochat} expands the image encoder and empowers LLMs to understand the visual content within videos via joint training. Video-ChatGPT~\cite{Video-chatgpt} employs a simple method of average pooling the frame-level tokens across both spatial and temporal dimensions to obtain a video-level tokens. Similarly, PLLaVA~\cite{pllava} trains a projection layer and utilizes an adaptive pooling layer to process frame-level tokens into video-level tokens. Following the BLIP-2 strategy, Video-LLaMA~\cite{video-llama} further employs pre-trained models such as ViT~\cite{evavit}, ImageBind~\cite{imagebind}, and LLaMA~\cite{llama}, executing cross-modal bootstrap training within the video\&audio domain. Video-LLaVA~\cite{video-llava} utilizes a pre-aligned encoder to achieve shared projections that can adapt to images and videos, consequently promoting synergistic training across image and video data. However, while the aforementioned models demonstrate excellent performance in short video understanding, their capabilities for comprehending long videos appear to be somewhat limited.

\subsection{Long Video Understanding Models}
As the number of long videos in the internet continues to grow, the demand for long video understanding models becomes increasingly pressing. Compared to short videos, long videos have a longer duration and contain diverse information, making long video understanding a challenging task. MIST~\cite{mist} introduces cascading segment and region selection modules, aiming to understand long video content while enhancing computational efficiency. In an effort to reduce the number of input frames for long videos, some researches~\cite{adaframe,scsampler} design various sparse video sampling techniques, retaining only the highlighted video frame content. Additionally, some works like MovieChat~\cite{moviechat} and MA-LMM~\cite{ma-lmm} integrate the global semantics of videos by devising memory bank mechanisms with token compression to store long term content information. However, as the number of frames in long videos continues to increase, their methods constantly compress tokens, which can result in various types of information being blended together, leading to information redundancy in the memory bank. To address the aforementioned issues, we propose an adaptive segmentation strategy for video sequences, employing local memory banks to store distinct events and temporally modeling each event through a global memory bank. This strategy aids in mitigating the information redundancy caused by coarse compression, ultimately augmenting the MLLM's capacity to comprehend the semantics of long-term videos.

\begin{figure*}[t]
  \centering
  \includegraphics[width=16cm,height=9cm]{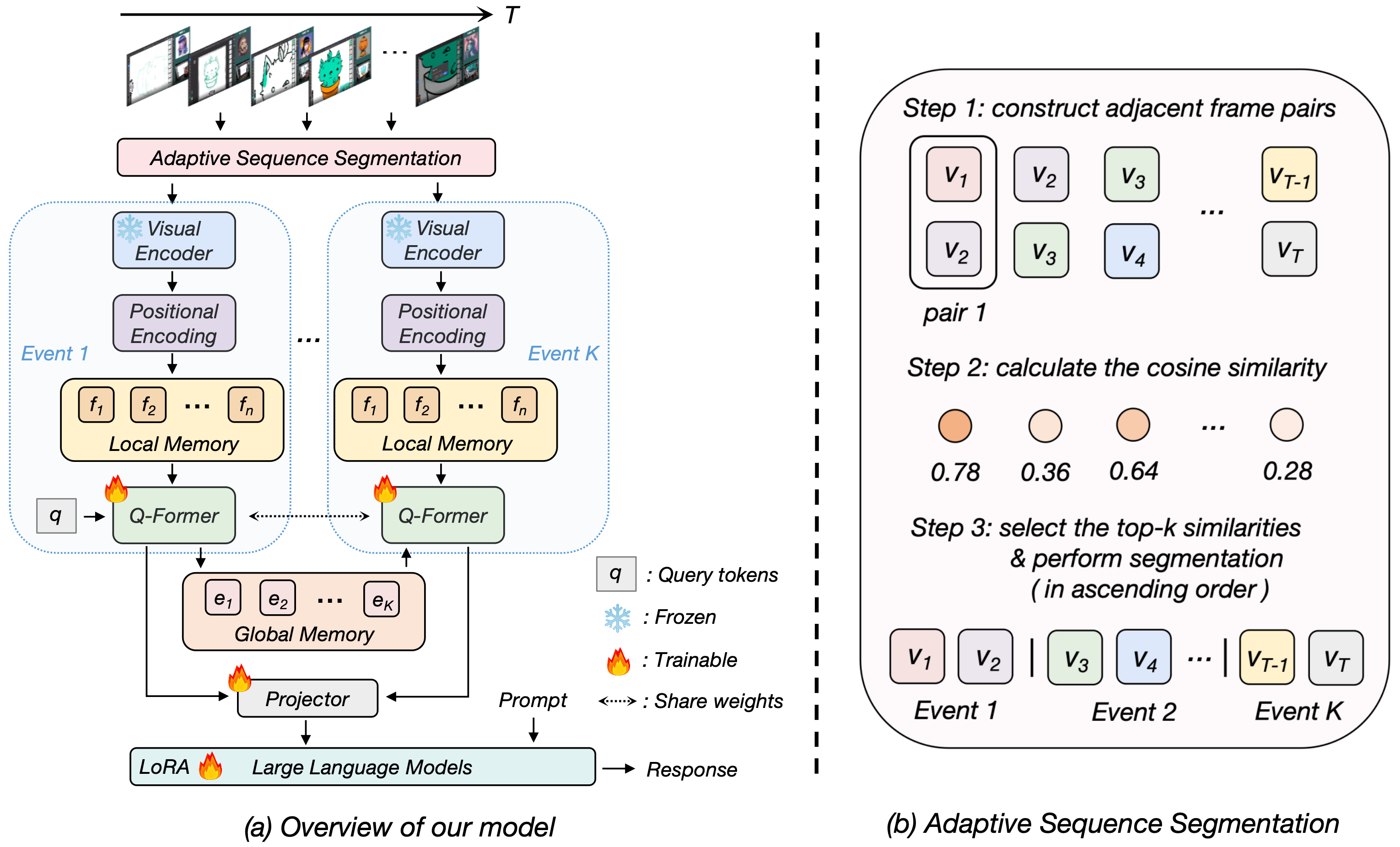}
  \caption{(a) Overview of the HEM-LLM. We first sequentially sample video frames and perform adaptive sequence segmentation to divide them into individual events. Then, we introduce local memory and global memory to model the temporal context for both intra-event and inter-event scenarios. In this way, the HEM-LLM can progressively mine video semantic information at multiple granularities to enhance multimodal understanding capabilities. Finally, we employ Q-Former to integrate and compress the visual tokens of each event, concatenate them, and form event-based visual tokens to be fed into the LLMs for text generation. (b) Adaptive Sequence Segmentation. To effectively and efficiently perform event-based adaptive segmentation, we proceed in three steps: (i) we establish pairwise adjacent frame pairs; (ii) we compute their token-level cosine similarities and select the minimum K-1 points as segmentation points; (iii) We split the video at the segmentation points to form K events.}
  \label{f2}
\end{figure*}

\section{Proposed methodology}
We propose a Hierarchical Event-based Memory-enhanced LLM for better understanding of long videos. We no longer compress the entire long video directly. Instead, we devise a novel adaptive sequence segmentation scheme to partition multiple events within the long video. Subsequently, local memory is utilized to establish intra-event contextual connections for each event separately. Then, to enhance the long-term inter-event dependencies, we introduce global memory to compress and inject historical event information, promoting the memory modeling of the current event. Finally, the well-modeled video tokens are inputted into the LLMs for video comprehension. The overall framework of the model is shown in Figure \ref{f2}.

\subsection{Adaptive Sequence Segmentation}
Essentially, video sequences consist of consecutive frames. Therefore, our model processes the videos in a sequential manner to facilitate temporal modeling of videos more conveniently. Meanwhile, compared to short videos, long videos encompass a more diverse array of scenes and event transitions. Capturing these transitions enables the accurate segmentation of different scenes and events within the video. The transition points between various scenes and events in a video are usually accompanied by significant changes in elements such as characters, backgrounds, activities, and shots. As a result, our model is designed to accurately and efficiently capture these significant changes.

Specifically, for a given video sequence \( V=\left \{ v_i \right \} _{i=1}^T\in \mathbb{R}^{3 \times T \times H \times W} \) of T frames, we employ token-level cosine similarity to capture significant changes between events and determine segmentation points. We first perform average pooling on the spatial dimensions of the T original frames containing RGB channels to integrate the global information of each frame. Next, we calculate the token-level cosine similarity \(Score\) between pairwise pooled frames to determine the magnitude of change between them. Finally, we apply \(TopK\) filtering to the similarity sequence, selecting the top K-1 frames as the segmentation points \(s_j\) between events in the video. Consequently, the video sequence \(V\) is adaptively partitioned into an event sequence \(E=\left \{ e_i \right \} _{i=1}^K \), where each event consists of a series of frames. The computation process is described below:
\begin{equation}
    \hat{V}=AdaptiveAvgPool(V)\in \mathbb{R}^{3 \times T}
\end{equation} 
where \(AdaptiveAvgPool\) represents AdaptiveAvgPool2d to perform average pooling of spatial dimensional information, namely \((H,W)\to (1,1)\).
\begin{equation}
    Score_i=cos(\hat{v}_i , \hat{v}_{i+1}) , i\in \left [ 1 , T-1 \right ] 
\end{equation}
where \(cos\) denotes token-level cosine similarity calculation.
\begin{equation}
    s_j=TopK_{min}(Score),j\in \left [ 1,K-1 \right ]
\end{equation} 
where \(TopK_{min}\) indicates taking the minimum \(K-1\) points in the sequence. We treat \(s_j\) as the segmentation point of the video \(V\), whereby \(V\) is divided into \(K\) events \(E\).

\subsection{Intra-event Local Memory Modeling}
For the event sequence \(E=\left \{ e_i \right \} _{i=1}^K \) obtained from the above computation, we perform independent local memory modeling for each event. We utilize a pre-trained visual encoder to perform token extraction for each frame in the event \(e_k\), yielding \(e_k=\left \{ f_i \right \}_{i=1}^n \in  \mathbb{R}^{d \times n \times p}\), \(p\) is the number of visual tokens per frame. To emphasize the temporal information between frames within an event, we assign frame-level positional encoding to each event respectively, thereby incorporating timestamp information into each frame. The main process can be written as:
\begin{equation}
    e_k=\left \{ f_1 \right \}_{i=1}^n \in  \mathbb{R}^{d \times n \times p} , f_i =VE(v_i)
\end{equation}
where \(e_k\) means the k-th event, \(VE\) represents the pre-trained visual encoder and \(p\) is the number of visual tokens per frame.

We perform positional encoding of \(f_i\) to add timestamp information as:
\begin{equation}
    f_i =f_i+PE(t_i)
\end{equation}
where \(PE\) denotes the frame-level positional encoding and \(t_i\) indicates the timestamp information of each frame \(f_i\).

Simultaneously, in order to effectively and efficiently bridge the semantic gap between video and text, we incorporate local memory into the Q-Former architecture proposed by BLIP-2, which uses a finite set of query tokens \(Q=\left \{ q_i \right \}_{i=1}^{32} \in  \mathbb{R}^{d \times 32}\) to facilitate multimodal representation learning. In this, we efficiently store the historical frame information in local memory in a concatenated manner to form a memory sequence \(LM=Concat\left [ f_1,f_2,...,f_n \right ] \in \mathbb{R}^{d \times np} \), which enables query tokens to fully learn the contextual information of the event. The Q-Former comprises two components: the self-attention among query tokens and the cross-attention between visual tokens and query tokens. 

The calculation process for the self-attention component among query tokens is as follows:
\begin{equation}
\label{e6}
    O=SelfAttn(Q_q,\hat{Q}_k,\hat{Q}_v)
\end{equation}
where \(O \in \mathbb{R}^{d \times 32}\) represents the output of the self-attention computation between query tokens. Inspired by \cite{ma-lmm}, we follow the \(LM\) operation to collect the query tokens \(\hat{Q}=Concat\left [ Q_1,Q_2,...,Q_n \right ] \in \mathbb{R}^{d \times 32n}\) at each time step for complementing the temporal information between the frames of the video sequence.

The cross-attention component with local memory is described as follows:
\begin{equation}
    O_c=CrossAttn(O_q,LM_k,LM_v)
\end{equation} 
where \(O_c \in \mathbb{R}^{d \times 32}\) denotes the final output of the Q-Former. We incorporate the \(O_c\) of each event as visual tokens into LLMs. 

\subsection{Inter-event Global Memory Modeling}
Since long videos contain more than one event and there are temporal relationships between consecutive events, it is essential to capture the inter-event long-term dependencies for a comprehensive understanding of video semantics. In view of this, after completing the local memory modeling within each event, we incorporate the events into the global memory to better learn the global information of the video. This process is shown below:
\begin{equation}
GM=Concat\left [ \hat{Q}_{1},\hat{Q}_{2},...,\hat{Q}_{N} \right ]  
\end{equation} 
where \(GM\) represents the global memory sequence, \(N\) denotes the number of events processed by the model up to the current event. Considering the limitations of computational resources, we employ token compression from \cite{ma-lmm} to control the number of tokens in \(GM\).

At the same time, to explore the progressive relationship between events, we inject information from the previous event while modeling the local memory of the current event. Specifically, we adopt \(GM\) to replace \(\hat{Q}\) in Equation (\ref{e6}) to incorporate historical events into the modeling of the current event. Thus, Equation (\ref{e6}) can be reformulated as:
\begin{equation}
    O=SelfAttn(Q_q,GM_k,GM_v)
\end{equation} 
where \(O \in \mathbb{R}^{d \times 32}\) denotes the output of the self-attention among query tokens.

\subsection{Text Tokens Generation}
Based on the calculated visual tokens \(O_c\) for each event, we concatenate them together as the visual tokens \(Z_v\) for the entire long video and integrate them into the LLMs through a projection layer. Subsequently, the LLMs perform auto-regressive reasoning on the basis of instruction prompts, ultimately generating high-quality text tokens \(Z_t\). Consequently, our model can achieve efficient video understanding with a smaller number of tokens, without being constrained by the LLM's context input and computational resource limitations. Furthermore, we apply cross entropy loss to the reasoning-generated tokens for Self-Supervised Fine-Tuning (SFT). The supervision process is as follows:
\begin{equation}
    Z_v=Concat\left [ O_{c1},O_{c2},...,O_{cK} \right ] \in \mathbb{R}^{d \times 32K}
\end{equation} 
where \(K\) indicates the number of events in the video.
\begin{equation}
    Z_t=LLM( \Delta (Z_v,Prompts))
\end{equation} 
where \(\Delta\) represents the projection layer.
\begin{equation}
    \mathcal L=CrossEntropy(Z_t,Target)
\end{equation} 
where \(CrossEntropy\) denotes the cross entropy loss function and \(Target\) means the target tokens.

\begin{table}[]
\centering
\caption{Quantitative evaluation for VQA task. The * represents the results reproduced in our environment. The best performance values are highlighted by \textbf{bold}.}
\label{t1}
\resizebox{0.47\textwidth}{!}{%
\begin{tabular}{@{}lcccccc@{}}
\toprule
\textbf{Method}                             & \multicolumn{2}{c}{\textbf{MSRVTT-QA}}                                 & \multicolumn{2}{c}{\textbf{MSVD-QA}}                                   & \multicolumn{2}{c}{\textbf{ActivityNet-QA}}                            \\ \midrule
\textit{\textbf{Exact matching :}} & \multicolumn{1}{c}{Top-1} & \multicolumn{1}{c}{Top-5} & \multicolumn{1}{c}{Top-1} & \multicolumn{1}{c}{Top-5} & \multicolumn{1}{c}{Top-1} & \multicolumn{1}{c}{Top-5} \\ \midrule
Flamingo-80B                             & 17.4                           & -                             & 35.6                          & -                             & -                             & -                             \\
BLIP-2                             & 9.2                           & -                             & 18.3                          & -                             & -                             & -                             \\
FrozenBiLM                             & 16.8                           & -                             & 32.2                          & -                             & -                             & -                             \\
mPLUG-2                            & 48.0                          & -                             & 58.1                          & -                             & -                             & -                             \\
UMT-L                              & 47.1                          & -                             & 55.2                          & -                             & 47.9                          & -                             \\
mPLUG-Owl2                         & 23.6                          & -                             & 42.4                          & -                             & -                             & -                             \\
InstructBLIP                       & 22.1                          & -                             & 41.8                          & -                             & -                             & -                             \\
Video-LLaMA                        & 46.5                          & -                             & 58.3                          & -                             & 45.5                          & -                             \\
MA-LMM*                            & 48.3                          & 75.4                          & 60.5                          & 84.0                          & 48.8                          & 69.5                          \\ \midrule
\textbf{Ours}                      & \textbf{49.8}                 & \textbf{77.0}                 & \textbf{61.5}                 & \textbf{85.8}                 & \textbf{50.6}                 & \textbf{71.1}                 \\ \midrule \midrule
\textit{\textbf{GPT3.5-assisted :}}   & Acc.                      & Score                         & Acc.                      & Score                         & Acc.                      & Score                         \\ \midrule
Video-LLaMA                        & 29.6                          & 1.8                           & 51.6                          & 2.5                           & 12.4                          & 1.1                           \\
VideoChat                          & 45.0                          & 2.5                           & 56.3                          & 2.8                           & 26.5                          & 2.2                           \\
Video-ChatGPT                      & 49.3                          & 2.8                           & 64.9                          & 3.3                           & 35.2                          & 2.7                           \\
Video-LLaVA                        & 59.2                          & \textbf{3.5}                  & 70.7                          & 3.9                           & 45.3                          & 3.3                           \\
mPLUG-Owl2                       & 46.7                             & 2.9                             & 65.4                          & 3.5                           & -                          & -                           \\
LLaMA-Adapter                       & 43.8                             & 2.7                             & 54.9                          & 3.1                           & -                          & -                           \\
Video-LLaMA2                       & -                             & -                             & 70.9                          & 3.8                           & 50.2                          & 3.3                           \\
LLaMA-VID                       & 57.7                             & 3.2                             & 69.7                          & 3.7                           & 47.4                          & 3.3                           \\
VideoChat2                         & 54.1                          & 3.3                           & 70.0                          & 3.9                           & 49.1                          & 3.3                           \\
VISTA-LLaMA                        & \textbf{60.5}                 & 3.3                           & 65.3                          & 3.6                           & 48.3                          & 3.3                           \\
LLaVA-NeXT                   & -                             & -                             & 67.8                          & 3.5                           & 53.5                          & 3.2                           \\ \midrule
\textbf{Ours}                      & 57.7                          & 3.3                           & \textbf{71.7}                 & \textbf{4.0}                  & \textbf{56.6}                 & \textbf{3.6}                  \\ \bottomrule
\end{tabular}%
}
\end{table}

\section{Experiments}
\subsection{Datasets and Evaluation}
To comprehensively validate the effectiveness of our proposed model, we conduct numerous experiments on video understanding downstream tasks such as video question answering, video captioning, and long video activity classification. Additionally, we employ multiple datasets with varying video lengths to verify the universality and generalization capabilities of our model.

\noindent \textbf{Video Question Answering.} In this task, we conduct experiments using four datasets of varying lengths. MSVD-QA~\cite{msrvttqa} and MSRVTT-QA~\cite{msrvttqa} are two diverse datasets that encompass a wide range of everyday life. Their videos have an average duration of 10-15 seconds. ActivityNet-QA~\cite{activitynetqa} is a large-scale dataset focused on human activities, with its videos having an average duration of 3 minutes. MovieChat-1K~\cite{moviechat} is a movie-related video understanding dataset, with an average duration of approximately 8 minutes. For the VQA task, we employ two evaluation methods to report experimental performance: (1) Exact matching, commonly used for assessing traditional VQA tasks. (2) GPT-assisted evaluation, which leverages the powerful capabilities of LLMs to measure the accuracy of model predictions and provides a relative score on a scale of 1-5.

\noindent \textbf{Video Captioning.} We evaluate our experiments using three publicly available datasets: MSVD~\cite{msvd}, MSRVTT~\cite{msrvtt}, and Youcook2~\cite{youcook2}. Among them, Youcook2 is a large-scale cooking tutorial dataset with an average video duration of 5.26 minutes. For this task, we report the METEOR~\cite{meteor} and CIDEr~\cite{cider} evaluation metrics.

\noindent \textbf{Long Video Activity Classification.} In this downstream task, we conduct experiments using two popular datasets, namely, Breakfast~\cite{breakfast} and COIN~\cite{coin}. Specifically, Breakfast contains a series of videos related to breakfast preparation, with an average length of 2.7 minutes. COIN is a large-scale video analysis dataset focusing on human activities. It collects videos from YouTube, covering 12 domains of everyday life, with an average video length of 2.36 minutes. For the classification task, we report the Top-1 and Top-5 accuracy.

\subsection{Implementation Details}
In terms of the visual backbone model, we adopt ViT-G/14~\cite{vit} from EVA-CLIP~\cite{evavit} to extract visual tokens, and we utilize Qformer from InstructBLIP~\cite{instructblip} to integrate and compress visual tokens. In the language model aspect, we employ Vicuna-7B~\cite{vicuna} for text reasoning. All experiments are conducted on 8 ATN 910B NPUs.

\begin{table}[]
\centering
\caption{Performance comparisons on MovieChat-1K \textit{Global Mode}. The \textit{\# Frames} denotes the number of video frames extracted by the model. The best performance values are highlighted by \textbf{bold}.}
\label{t2}
\resizebox{0.47\textwidth}{!}{%
\begin{tabular}{@{}lcccc@{}}
\toprule
\textbf{Method}        & \textbf{Text Decoder}       & \textbf{\# Frames}    & \textbf{Accuracy}      & \textbf{Score}         \\ \midrule
GIT           & non-LLM based      & 6            & 28.8          & 1.83          \\
mPLUG-2       & non-LLM based      & 8            & 31.7          & 2.13          \\ \midrule
VideoChat     & LLM based          & 32           & 57.8          & 3.00          \\
Video-LLaMA   & LLM based          & 32           & 51.7          & 2.67          \\
Video-ChatGPT  & LLM based          & 100          & 47.6          & 2.55          \\
MovieChat     & LLM based          & 2048         & 62.3          & 3.23          \\
MovieChat+    & LLM based          & 2048         & 71.2          & 3.51          \\ \midrule
\textbf{Ours} & \textbf{LLM based} & \textbf{100} & \textbf{90.6} & \textbf{4.46} \\ \bottomrule
\end{tabular}
}
\end{table}

\begin{table}[]
\centering
\caption{Performance comparisons. The M indicates METEOR metric and C represents CIDEr metric. The * represents the results reproduced in our environment. The best performance values are highlighted by \textbf{bold}.}
\label{t3}
\resizebox{0.45\textwidth}{!}{%
\begin{tabular}{@{}lcccccc@{}}
\toprule
\multirow{2}{*}{\textbf{Method}} & \multicolumn{2}{c}{\textbf{MSRVTT}}    & \multicolumn{2}{c}{\textbf{MSVD}}       & \multicolumn{2}{c}{\textbf{YouCook2}}   \\ \cmidrule(l){2-7} 
                        & M        & C         & M        & C          & M        & C          \\ \midrule
UniVL                   & 28.2          & 49.9          & 29.3          & 52.8           & -             & 127.0          \\
SwinBERT                & 29.9          & 53.8          & 41.3          & 120.6          & 15.6          & 109.0          \\
GIT                     & 32.9          & 73.9          & \textbf{51.1} & 180.2          & 17.3          & 129.8          \\
VideoCoca               & -             & 73.2          & -             & -              & -             & 128.0          \\
Video-LLaMA             & 32.9          & 71.6          & 49.8          & 175.3          & 16.5          & 123.7          \\
MA-LMM*                 & 32.2          & 73.8          & 49.7          & 178.7          & 16.8          & 127.3          \\ \midrule
\textbf{Ours}           & \textbf{33.0} & \textbf{76.5} & 49.8          & \textbf{181.8} & \textbf{17.7} & \textbf{137.2} \\ \bottomrule
\end{tabular}%
}
\end{table}

\begin{table}[]
\centering
\caption{Performance comparisons. The * represents the results reproduced in our environment. We report the Top-1 and Top-5 accuracy. The best performance values are highlighted by \textbf{bold}.}
\label{t4}
\small
\resizebox{0.4\textwidth}{!}{%
\begin{tabular}{@{}lcccc@{}}
\toprule
\multirow{2}{*}{\textbf{Method}} & \multicolumn{2}{c}{\textbf{Breakfast}} & \multicolumn{2}{c}{\textbf{COIN}}      \\ \cmidrule(l){2-5} 
                        & Top-1        & Top-5        & Top-1        & Top-5        \\ \midrule
TSN             & -          & -             & 73.4             & -             \\  
VideoGraph      & 69.5          & -             & -             & -             \\  
Timeception             & 71.3          & -             & -             & -             \\
GHRM                    & 75.5          & -             & \textbf{-}    & -             \\
D-Sprv                  & 89.9          & -             & 90.0          & -             \\
ViS4mer                 & 88.2          & -             & 88.4          & -             \\
TranS4mer               & 90.3          & -             & 89.3          & -             \\
S5                      & 90.7          & -             & 90.8          & -             \\
MA-LMM*                 & 91.8          & \textbf{99.4} & 92.6          & 96.6          \\ \midrule
\textbf{Ours}           & \textbf{95.8} & 99.2          & \textbf{94.4} & \textbf{98.3} \\ \bottomrule
\end{tabular}%
}
\end{table}

\subsection{Experimental Results}
\noindent \textbf{Video Question Answering.} To conduct a comprehensive comparison with existing models, we perform extensive experiments on four open-ended datasets employing two evaluation ways. The results are shown in Table \ref{t1} and Table \ref{t2}. On the MSRVTT-QA, MSVD-QA, and ActivityNet-QA datasets, our model demonstrates superior performance compared to existing models~\cite{FrozenBiLM,umt-l,llama-adapter} under both evaluation methods. For instance, our model achieves gains of 3.6\%, 1.7\%, and 7.5\% on the three datasets, respectively, compared to VideoChat2~\cite{videochat2}. This indicates that our model can more accurately extract the semantic information from videos to enhance its understanding capabilities. On the MovieChat-1K, our model achieves a 19.4\% improvement over MovieChat+~\cite{moviechat+} when only sampling 100 frames, further validating the effectiveness of our model.

Remarkably, as the video duration increases, the gains achieved by our model also become more substantial, indicating that our model has good generalization for a wide range of video lengths. At the same time, this also validates that our model can accurately segment the individual events in long videos and effectively establish the intra-event and inter-event temporal context relationships.

\noindent \textbf{Video Captioning.} To further validate the text generation capabilities of our model, we perform extensive experiments on three classic video captioning datasets, namely MSVD, MSRVTT, and YouCook2. The experimental results can be found in Table \ref{t3}. Our model outperforms existing models~\cite{git,videococa,swinbert,univl} on both the METEOR and CIDEr evaluation metrics. Moreover, compared to the recent VideoLLaMA~\cite{video-llama} and MA-LMM~\cite{ma-lmm}, which also adopt LLMs as their language reasoning models, our model significantly surpasses them on all evaluation metrics. These results demonstrate that our model possesses strong video understanding abilities and can generate high-quality captions.

\noindent \textbf{Long Video Activity Classification.} For classification tasks, instead of having the model predict scores within a closed set of class labels, we make the model generate open-ended labels, which closely aligns with the practical requirements of real-world applications. We conduct numerous experiments on Breakfast and COIN datasets, and the results are presented in Table \ref{t4}. The results show that our model achieves a 4\% and 1.8\% improvement in the Top-1 accuracy metric compared to the existing the state-of-the-art model on the two datasets, respectively. Compared to other models~\cite{s5,timeception,d-sprv,ghrm,tsn,videograph}, our model also has significant advantages. This not only fully validates the effectiveness of our model but also highlights the importance of segmenting multiple events in long videos and establishing their long-term dependency relationships.

\begin{table}[]
\centering
\caption{The results of the ablation study on Breakfast. The best performance values are highlighted by \textbf{bold}.}
\label{t5}
\resizebox{0.43\textwidth}{!}{%
\begin{tabular}{ccccc}
\toprule
\textbf{Model}               & \multicolumn{1}{c}{\textbf{Local}} & \multicolumn{1}{c}{\textbf{Global}} & \multicolumn{1}{c}{\textbf{ASS}}  & \multicolumn{1}{c}{\textbf{Top-1}} \\ \midrule
\multicolumn{1}{c}{1} & \XSolidBrush                                  & \XSolidBrush                                  & \XSolidBrush                                                                     & 74.6                             \\
2                            & \Checkmark                                 & \XSolidBrush                                  & \XSolidBrush                                                                     & 91.0                             \\
3                            & \Checkmark                                 & \Checkmark                                 & \XSolidBrush                                                                     & 94.1                             \\
4                            & \Checkmark                                 & \Checkmark                                 & \Checkmark                                                                    & \textbf{95.8}                             \\
 \bottomrule
\end{tabular}%
}
\end{table}

\begin{table}[]
\centering
\caption{Results of calculating cosine similarity using different forms of tokens. The top-2 performance values are highlighted by \textbf{bold} and \underline{underline}, respectively.}
\label{t6}
\resizebox{0.45\textwidth}{!}{%
\begin{tabular}{@{}lcc@{}}
\toprule
\multirow{2}{*}{\textbf{Method}} & \textbf{Breakfast}     & \textbf{MSVD-QA}       \\ \cmidrule(l){2-3} 
                        & Top-1         & Top-1         \\ \midrule
ViT Tokens+Avgpool      & 92.4          & 61.2          \\
ViT CLS Token           & 91.3          & {\underline{61.3}}    \\
Original Image+Avgpool  & \textbf{95.8} & \textbf{61.5} \\ \midrule
Linear+Sigmoid          & {\underline{94.7}}    & 60.9          \\ \bottomrule
\end{tabular}%
}
\end{table}

\begin{figure}[t]
  \centering
  \includegraphics[width=.48\textwidth]{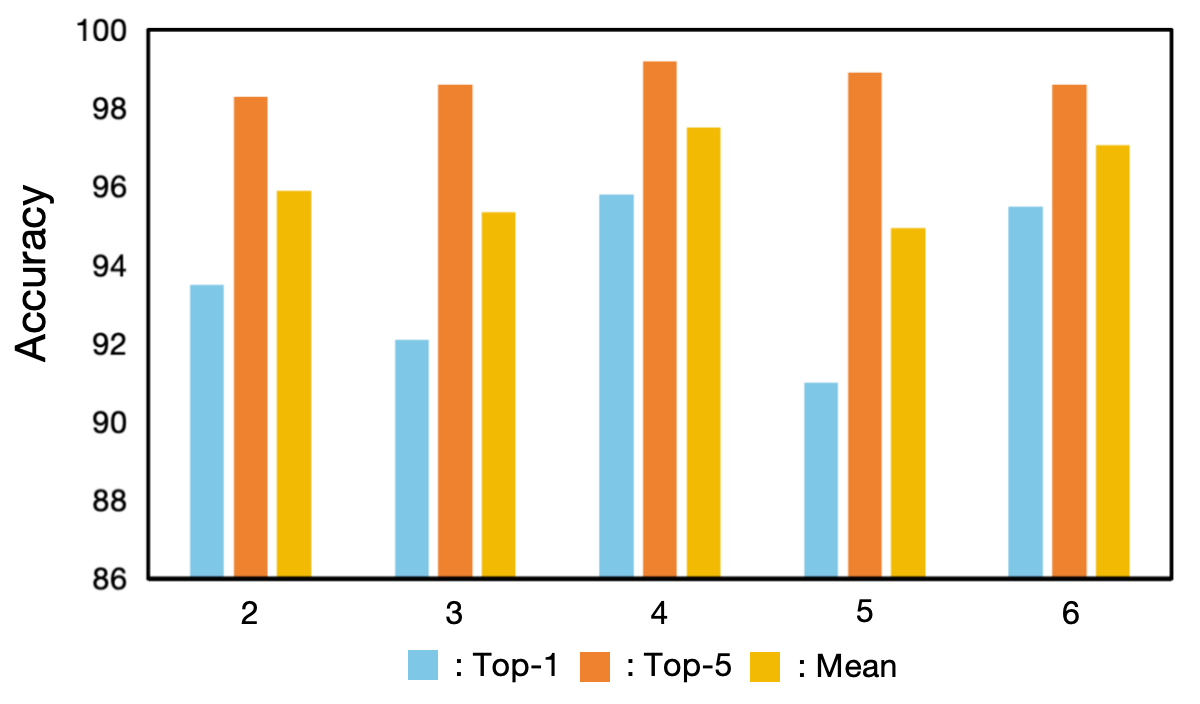}
  \caption{The study on the number of event segments. The \(Mean\) represents the average of Top-1 and Top-5 accuracy.}
  \label{f3}
\end{figure}

\subsection{Ablation Studies}
In this section, we conduct a comprehensive ablation analysis of the various components of HEM-LLM to ensure the completeness of the experiments. We perform extensive experiments on the Breakfast and MSVD-QA datasets and report Top-1 and Top-5 accuracy as research metrics.

\noindent \textbf{Analysis of the validity of individual components.} We perform an ablation analysis of the individual components proposed in this paper. The experimental results are shown in Table \ref{t5}, where Local represents Intra-event Local Memory Modeling, Global denotes Inter-event Global Memory Modeling, and ASS means Adaptive Sequence Segmentation. The results demonstrate that as different components are successively incorporated, the Top-1 accuracy metric continuously improves, showing a positive correlation with the addition of different components. This validates the effectiveness of the proposed components and highlights the value and necessity of event segmentation and multi-granular memory modeling for long videos.

\noindent \textbf{Investigation of calculating cosine similarity using different forms of tokens.} During the adaptive sequence segmentation process in the model, we investigate the form of tokens used to calculate cosine similarity. We adopt the Top-1 and Top-5 accuracy on the Breakfast as reference metrics. The experiment results are shown in Table \ref{t6}. In table \ref{t6}, Original Image+Avgpool represents visual tokens formed by applying average pooling on the spatial dimensions of the original image with RGB channels. ViT Tokens+Avgpool implies that the original image is first pre-processed with ViT for token extraction and then average pooling is applied. ViT CLS Token denotes the CLS Token encoded by the ViT model. As mentioned above, cosine similarity is used to obtain segmentation points. Furthermore, we also explore a method for the model to autonomously predict segmentation points, i.e., using a Linear+Sigmoid structure to predict segmentation points in the video sequence. The experimental results indicate that employing Original Image+Avgpool to form visual tokens and subsequently calculating segmentation points yields the best performance.

\noindent \textbf{Study on the number of event segments.} To investigate the impact of different event segmentation quantities on experimental performance, we conduct a quantitative study with the number of events ranging from 2 to 6, with a step size of 1. We perform experiments on the Breakfast, and the results are shown in Figure \ref{f3}. We report three performance metrics, namely Top-1, Top-5, and their average. The results show that the performance is optimal when the number of event segments is 4. Therefore, we set the number of event segments on the Breakfast dataset to 4.

\begin{figure*}[t]
  \centering
  \includegraphics[width=1.0\textwidth]{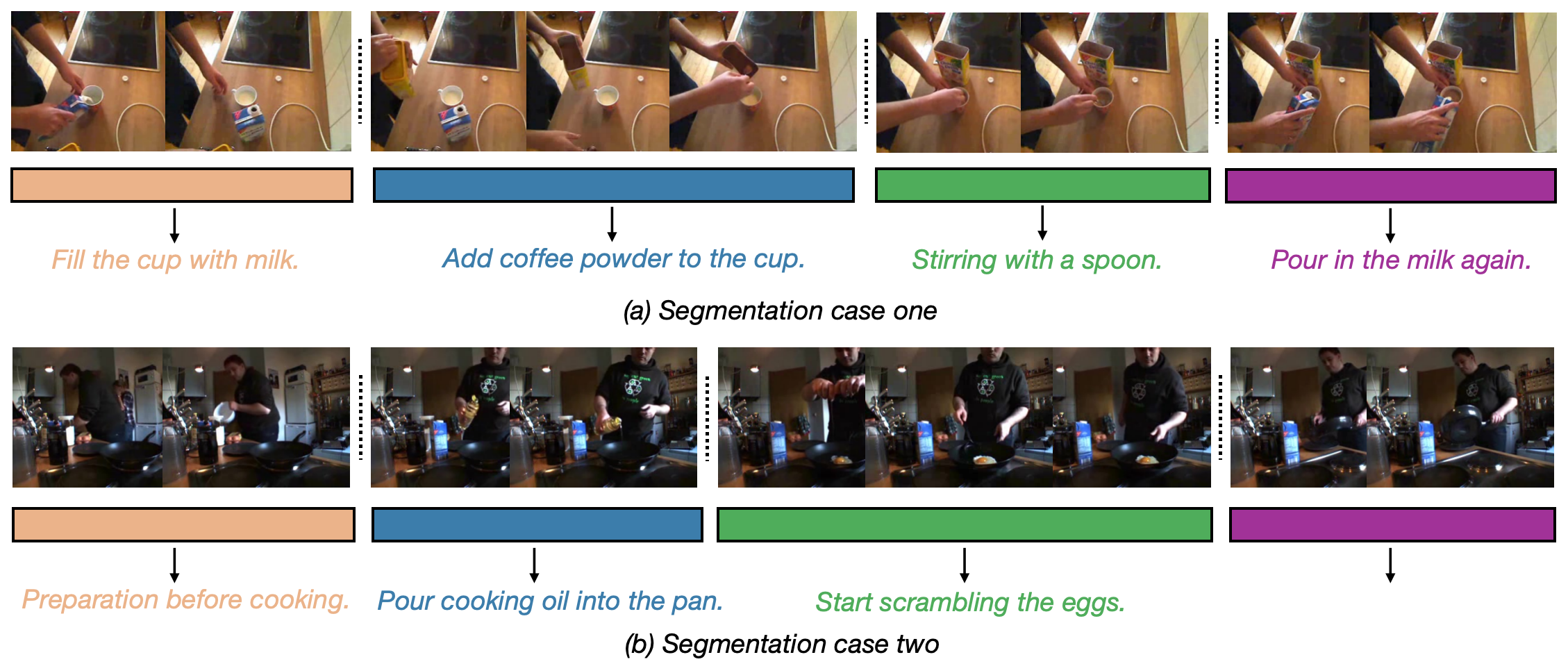}
  \caption{Two cases of event segmentation on the Breakfast.}
  \label{f4}
\end{figure*}

\begin{figure*}[t]
  \centering
  \includegraphics[width=1.0\textwidth]{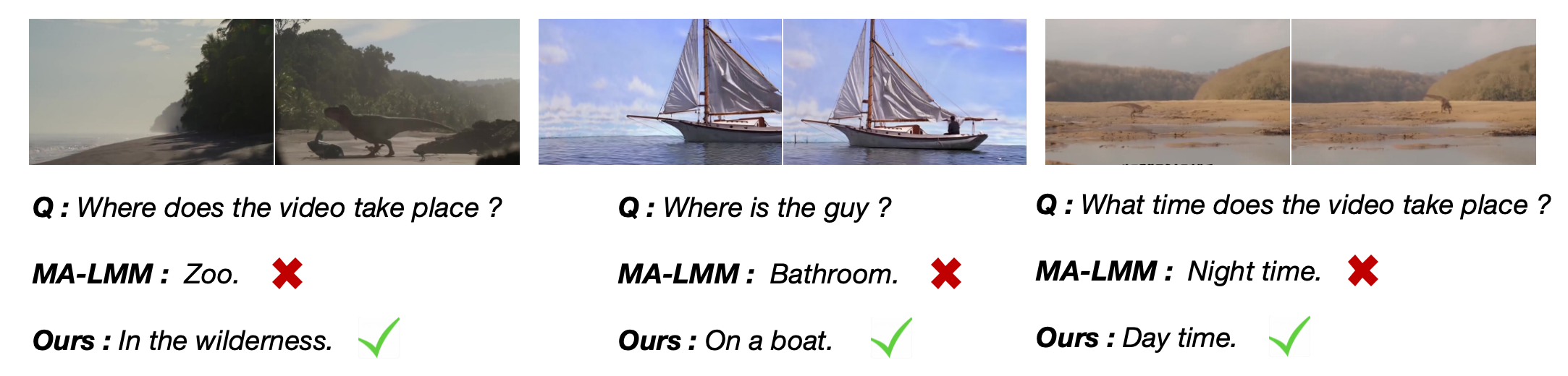}
  \caption{Qualitative analysis of video question answering on MovieChat-1K.}
  \label{f5}
\end{figure*}

\subsection{Visualization and Analysis}
To more intuitively validate the accuracy of our proposed adaptive sequence segmentation, we perform the case visualization analysis on the Breakfast. The visualization results are shown in Figure \ref{f4}. In (a), we determined three segmentation points based on the cosine similarity calculations between adjacent frame pairs. According to these points, the video can be divided into four event scenes: (i) fill the cup with milk, (ii) add coffee powder to the cup, (iii)stir with a spoon, and (iv) pour in the milk again. This demonstrates that our adaptive sequence segmentation can accurately segment videos to assist the model in mining video semantics at multiple granularities for better understanding of long videos. Similarly, (b) of Figure \ref{f4} also validates this, demonstrating that our method is valuable for long video modeling, as it reduces the likelihood of multiple events being mixed together.
Moreover, we also conduct a qualitative analysis of the model on the video question answering task to more comprehensively analyze its performance. We adopt MA-LMM as a comparison model and perform visualization experiments on the MovieChat-1K dataset. As shown in Figure ~\ref{f5}, the results demonstrate that our model exhibits advantages in time and location-related questions, thus proving its effectiveness. Compared to MA-LMM, which compresses the entire video, our model first segments events and then performs hierarchical memory modeling for intra-event and inter-event scenarios. This method reduces the impact of information redundancy and enhances the semantics of key information, thereby improving the understanding capabilities for long videos.

\section{Conclusion}
In this paper, we propose a Hierarchical Event-based Memory-enhanced Large Language Model (HEM-LLM) for better understanding of long videos. To learn the semantics of individual events in long videos more precisely, we no longer directly compress the visual tokens of the entire long video. Instead, we design a novel adaptive sequence segmentation scheme to segment multiple events in long videos. This allows our model to handle each event respectively, thus reducing information confusion. Subsequently, we conduct multi-granular memory modeling for intra-event and inter-event scenarios to establish long-term dependencies in long video sequences, thereby enhancing the video understanding capabilities of the model. Finally, we carry out extensive experiments on various benchmarks of multiple video understanding tasks, demonstrating the effectiveness and universality of our proposed model. 

\bibliography{aaai25}

\vspace{-0.3cm}
\section{Appendix}

\subsection{Investigation of the number of event segments.} 
\begin{figure}[H]
  \includegraphics[width=.46\textwidth]{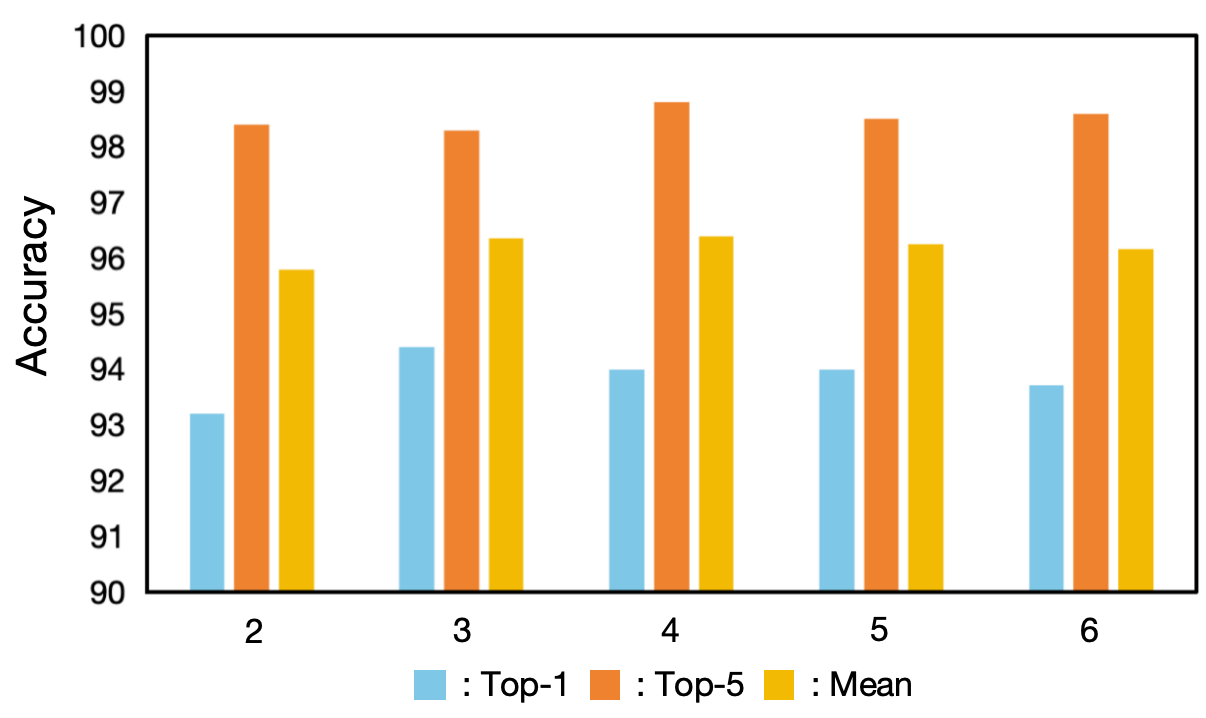}
  \caption{The study on the number of event segments. The \(Mean\) represents the average of Top-1 and Top-5 accuracy.}
  \label{a1}
\end{figure}

study the impact of different event segmentation quantities on experimental performance, we also perform experiments on the COIN, and the results are shown in Figure \ref{a1}. We also conduct a quantitative study with the number of events ranging from 2 to 6, with a step size of 1. We report three performance metrics, namely Top-1, Top-5, and their average. The results show that the performance is optimal when the number of event segments is 3. Therefore, we set the number of event segments on the COIN dataset to 3. Furthermore, for the VQA, we set the number of event segments to 2 across all datasets. For the video captioning, we set the number of event segments to 3 across all datasets.

\subsection{Study of different sizes of LLM.} We conduct experiments on YouCook2 using Vicuna with various parameters, as shown in Table \ref{at1}. The results demonstrate that our method outperforms existing models across different scales of LLM, which further validates the effectiveness and versatility of our proposed method. Remarkably, our method outperforms the previous methods which utilize 7B LLM when using a parameter count of only 3.7B LLM, further demonstrating the superiority of our method.

\begin{table}[H]
\centering
\caption{Results of different sizes of LLM. The F denotes frozen parameters and S stands for LoRA fine-tuning.}
\label{at1}
\resizebox{0.38\textwidth}{!}{%
\begin{tabular}{@{}ccccc@{}}
\toprule
\textbf{LLM}                     & \textbf{Params} & \textbf{F/S} & \textbf{METEOR}             & \textbf{CIDEr}              \\ \midrule
\multirow{2}{*}{Vicuna} & 3.7B   & F   & \textbf{17.8} & 130.7          \\
                        & 3.7B   & S   & 17.2          & 131.4          \\ \midrule
\multirow{2}{*}{Vicuna} & 5.5B   & F   & 17.4          & 132.9          \\
                        & 5.5B   & S   & 17.1          & 133.2          \\ \midrule
\multirow{2}{*}{Vicuna} & 7B     & F   & 17.2          & {\underline{133.4}}    \\
                        & 7B     & S   & {\underline{17.7}}    & \textbf{137.2} \\ \bottomrule
\end{tabular}%
}
\vspace{-0.4cm}
\end{table}

\subsection{Training Strategy}
During the Adaptive Sequence Segmentation process, the segmentation points for each video are distinct. Furthermore, a batch size greater than one leads to improved performance during training. Therefore, we devise two sampling schemes to enable a batch size larger than one during the training phase. The schemes are shown in Algorithm \ref{algorithm1} and Algorithm \ref{algorithm2}, where \(uniform\_sampling([0,i],S)\) denotes sampling \(S\) frames uniformly over the range \([0,i]\). 

\begin{algorithm}[] 
\caption{Sampling Frames Scheme}
\label{algorithm1}
\begin{algorithmic}[1] 
\STATE Let $K=2, K$ denotes the number of events. \#~[1]
\STATE Perform segmentation point $P$ calculations. \#~[B]
\STATE $S1=max(P)$ \#~[B]
\STATE $S2=max(T-P), T$ denotes the sum of frames. \#~[B]
\FOR{$i$ in $len(S1)$}
\STATE $list_1 = uniform\_sampling ([0,P[i]],S1_i)$
\STATE $image\_list_1.append(list_1)$
\ENDFOR
\STATE $image\_list.append(image\_list_1)$
\FOR{$j$ in $len(S2)$}
\STATE $list_2 = uniform\_sampling ([P[j],T],S2_j)$
\STATE $image\_list_2.append(list_2)$
\ENDFOR
\STATE $image\_list.append(image\_list_2)$
\STATE \textbf{return} $image\_list$
\end{algorithmic}
\end{algorithm}
\vspace{-0.5cm}

\begin{algorithm}[] 
\caption{Sampling Frames Scheme}
\label{algorithm2}
\begin{algorithmic}[1] 
\STATE Let $K=2, K$ denotes the number of events. \#~[1]
\STATE Perform segmentation point $P$ calculations. \#~[B]
\STATE $AF=Avg(P)$
\STATE $S1=AF$ \#~[B]
\STATE $S2=T-AF, T$ denotes the sum of frames. \#~[B]
\FOR{$i$ in $len(S1)$}
\STATE $list_1 = uniform\_sampling ([0,P[i]],AF)$
\STATE $image\_list_1.append(list_1)$
\ENDFOR
\STATE $image\_list.append(image\_list_1)$
\FOR{$j$ in $len(S2)$}
\STATE $list_2 = uniform\_sampling ([P[j],T],AF)$
\STATE $image\_list_2.append(list_2)$
\ENDFOR
\STATE $image\_list.append(image\_list_2)$
\STATE \textbf{return} $image\_list$
\end{algorithmic}
\end{algorithm}

We perform experiments on several datasets with the above two schemes and the results are shown in Table \ref{at2}. We report the Top-1 accuracy for each dataset. Considering the overall results, we adopt Algorithm \ref{algorithm1} for uniform sampling.

\begin{table}[H]
\centering
\caption{Performance comparisons. The top-2 performance values are highlighted by \textbf{bold} and \underline{underline}, respectively. }
\label{at2}
\resizebox{0.40\textwidth}{!}{%
\begin{tabular}{@{}lccc@{}}
\toprule
\textbf{Method}                     & \textbf{Scheme} & \textbf{Breakfast} & \textbf{COIN}                           \\ \midrule
\multirow{2}{*}{HEM-LLM} & 1  & \textbf{95.8}   & \textbf{94.4}           \\
                        & 2   & \underline{94.4}   & \underline{93.3}                    \\ \midrule 
\textbf{Method}                     & \textbf{Scheme} & \textbf{MSVD} & \textbf{MSRVTT}                           \\ \midrule                        
\multirow{2}{*}{HEM-LLM} & 1   & \underline{61.5}   & \textbf{49.8}                    \\
                        & 2   & \textbf{61.6}   & \textbf{49.8}                   \\ \midrule
\textbf{Method}                     & \textbf{Scheme} & \multicolumn{2}{c}{\textbf{ActivityNet}}  \\
\midrule
\multirow{2}{*}{HEM-LLM} & 1        & \multicolumn{2}{c}{\underline{50.6}}              \\
                        & 2        & \multicolumn{2}{c}{\textbf{50.8}}     \\ \bottomrule
\end{tabular}%
}
\end{table}

\subsection{More Visualization Cases}
\begin{figure}[H]
  \includegraphics[width=.46\textwidth]{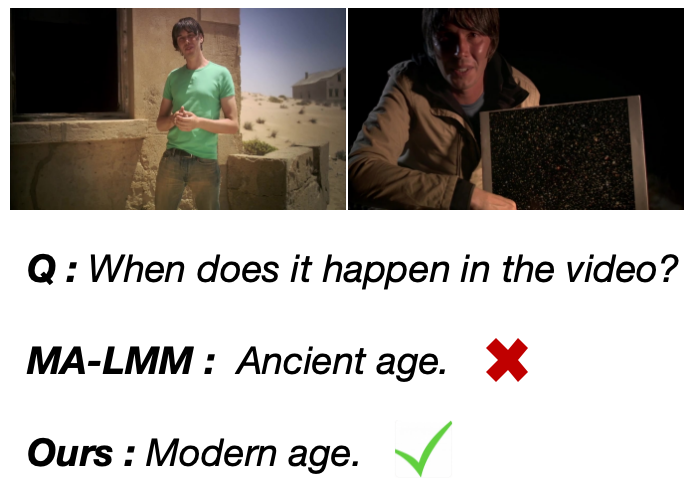}
  \caption{The case of VQA on the MovieChat-1K.}
  \label{a2}
\end{figure}

\begin{figure}[H]
  \includegraphics[width=.46\textwidth]{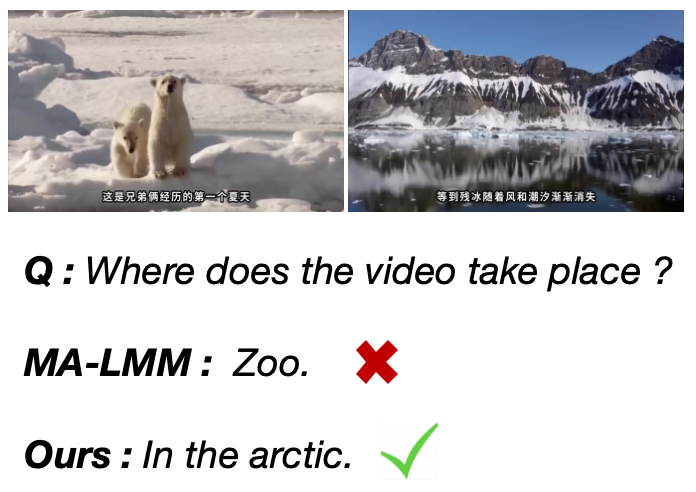}
  \caption{The case of VQA on the MovieChat-1K.}
  \label{a3}
\end{figure}

\begin{figure}[H]
  \includegraphics[width=.46\textwidth]{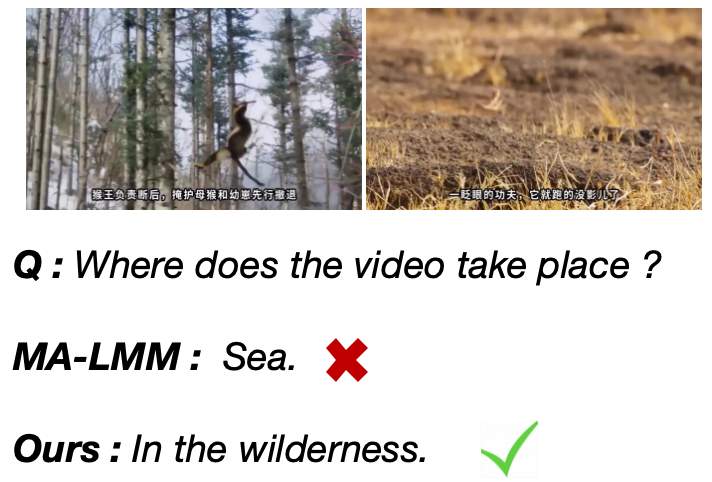}
  \caption{The case of VQA on the MovieChat-1K.}
  \label{a4}
\end{figure}

To further analyze the model's performance, we conduct more qualitative analyses on the video question answering task, as illustrated in Figures \ref{a2}, \ref{a3}, and \ref{a4}. The qualitative experimental results provide an intuitive demonstration of the advantages of our model compared to other model~\cite{ma-lmm}. The visualization results also corroborate the effectiveness and versatility of our model.

\subsection{Future works}
In the future, we will also extend our model to more datasets for video comprehension tasks. We will also explore other methods related to memory modeling, including token storage methods within the memory space, among others, to achieve more effective mining of context relationships.

\end{document}